\documentclass[10pt,twocolumn,letterpaper]{article}

\usepackage{cvpr}              

\usepackage{graphicx}
\usepackage{amsmath}
\usepackage{amssymb}
\usepackage{booktabs}

\usepackage{microtype} 

\makeatletter
\@namedef{ver@everyshi.sty}{} 
\makeatother

\graphicspath{{./graphics/}} 
\usepackage{animate} 
\usepackage{tabularx} 
\usepackage{booktabs} 
\usepackage{caption} 
\usepackage{cuted}
\usepackage{capt-of}
\usepackage{multirow}
\usepackage{units}
\usepackage{pifont}
\usepackage{colortbl}
\usepackage{enumitem}
\usepackage[accsupp]{axessibility}
\usepackage[toc,page]{appendix}
\usepackage[symbol]{footmisc}

\DeclareMathOperator*{\argmax}{argmax}
\newcommand{\ra}[1]{\renewcommand{\arraystretch}{#1}}
\newcommand{\cmark}{\ding{51}}

\definecolor{Gray}{gray}{0.95}
\newcolumntype{g}{>{\columncolor{Gray}}c}

\usepackage[pagebackref,breaklinks,colorlinks]{hyperref}

\def\cvprPaperID{8082}
\def\confName{CVPR}
\def\confYear{2022}

\begin{document}

\title{Layered Depth Refinement with Mask Guidance}

\author{
Soo Ye Kim\textsuperscript{1}
\qquad Jianming Zhang\textsuperscript{2} 
\qquad Simon Niklaus\textsuperscript{2} 
\qquad Yifei Fan\textsuperscript{2} \\
\qquad Simon Chen\textsuperscript{2} 
\qquad Zhe Lin\textsuperscript{2} 
\qquad Munchurl Kim\textsuperscript{1} 
\\
\begin{tabular}{*{2}{>{\centering}p{.3\textwidth}}}
\textsuperscript{1}KAIST, Republic of Korea & \textsuperscript{2}Adobe Inc., USA \tabularnewline
\end{tabular}
}

\maketitle

\begin{abstract}
Depth maps are used in a wide range of applications from 3D rendering to 2D image effects such as Bokeh. However, those predicted by single image depth estimation (SIDE) models often fail to capture isolated holes in objects and/or have inaccurate boundary regions. Meanwhile, high-quality masks are much easier to obtain, using commercial auto-masking tools or off-the-shelf methods of segmentation and matting or even by manual editing. Hence, in this paper, we formulate a novel problem of mask-guided depth refinement that utilizes a generic mask to refine the depth prediction of SIDE models. Our framework performs layered refinement and inpainting/outpainting, decomposing the depth map into two separate layers signified by the mask and the inverse mask. As datasets with both depth and mask annotations are scarce, we propose a self-supervised learning scheme that uses arbitrary masks and RGB-D datasets. We empirically show that our method is robust to different types of masks and initial depth predictions, accurately refining depth values in inner and outer mask boundary regions. We further analyze our model with an ablation study and demonstrate results on real applications. 
More information can be found on our project page.\footnote[2]{\url{https://sooyekim.github.io/MaskDepth/}}
\end{abstract}
\section{Introduction}

\begin{figure}
\centering
\includegraphics[width=\linewidth]{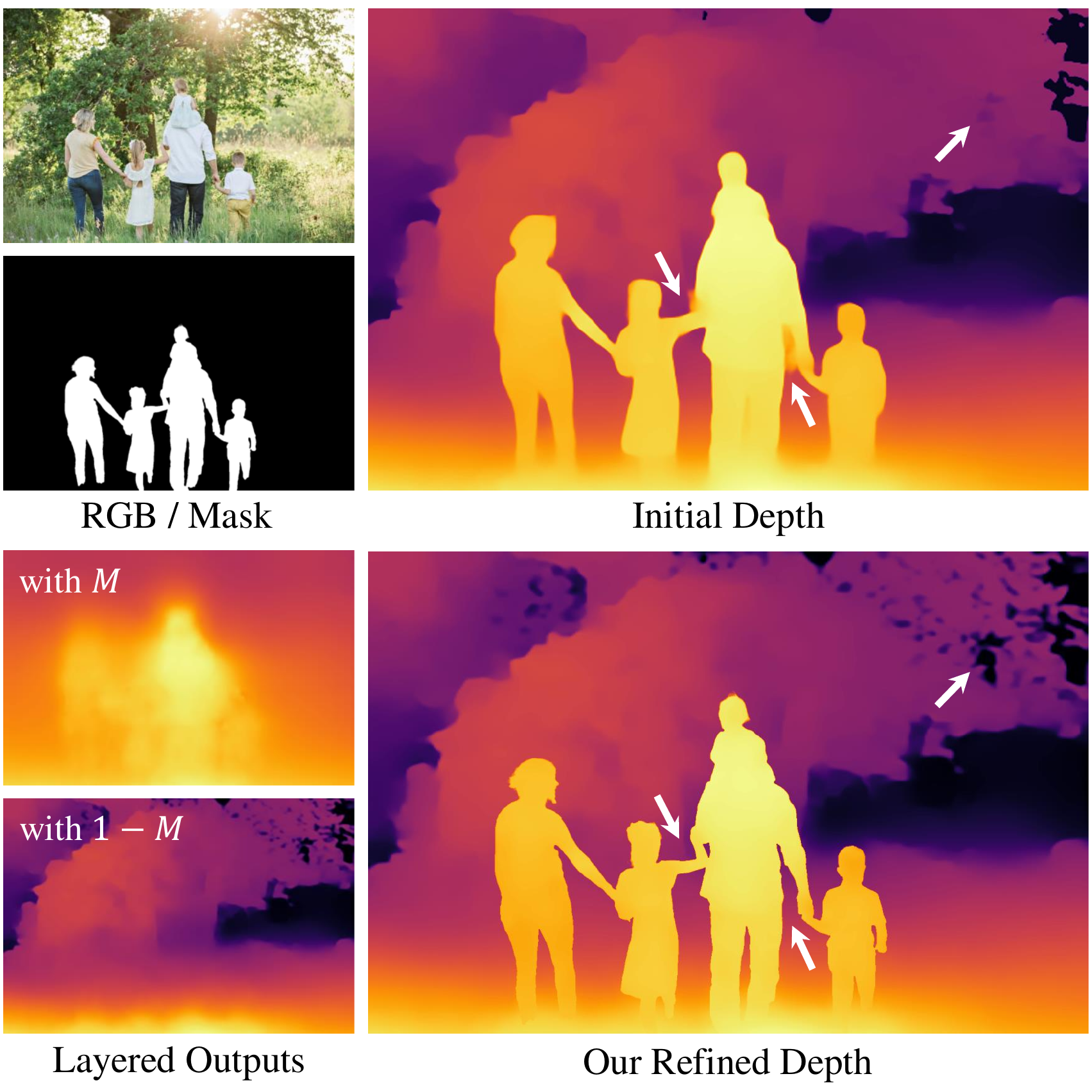}
\caption{Our layered depth refinement result on an initial prediction by DPT \cite{Ranftl_ICCV_2021}. Aided by a high-quality mask $M$ generated with an auto-masking tool \cite{removebg}, our method is able to accurately refine mask boundaries and correct depth values in isolated hole regions between body parts. Regions in $M$ and $1-M$ are refined and inpainted/outpainted separately with our layered approach.
}
\label{fig:teaser}
\vspace{-0.1em}
\end{figure}

Recent progress in deep learning has enabled the prediction of fairly reliable depth maps from single RGB images \cite{Li_CVPR_2018, Xian_2020_CVPR, Ranftl_TPAMI_2020, Ranftl_ICCV_2021}.
However, despite the specialized network architectures \cite{Fu_CVPR_2018, Qi_CVPR_2018, Ranftl_ICCV_2021} and training strategies \cite{Xian_CVPR_2018, Ranftl_TPAMI_2020} in single image depth estimation (SIDE) models, the estimated depth maps are still inadequate in the following aspects: (i) depth boundaries tend to be blurry and inaccurate; (ii) thin structures such as poles and wires are often missing; and (iii) depth values 
in narrow or isolated background regions (e.g., between body parts in humans) are often imprecise, as shown in the initial depth estimation in Figure~\ref{fig:teaser}. 
Addressing these issues within a single SIDE model can be very challenging due to limited model capacity and the lack of high-quality RGB-D datasets.

\begin{figure*}
\centering
\includegraphics[width=\linewidth]{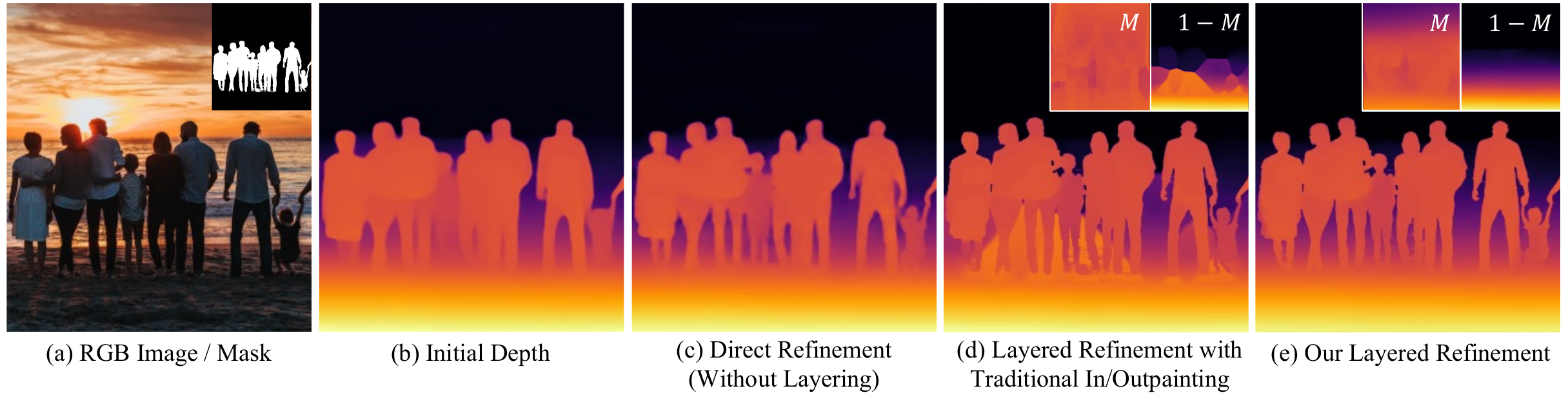}
\caption{Refined depth maps with the guidance of a high-quality mask. (b) The initial depth prediction \cite{Ranftl_ICCV_2021} has blurry boundaries and misses isolated hole regions between human body parts. (c) Direct refinement by training on a paired dataset \cite{Roberts_ICCV_2021} improves the initial depth but still has blurry boundaries. 
Layered refinement results in sharp edges due to the final compositing step using the mask, although (d) naive in/outpainting \cite{Telea_JGT_2004} generates artifacts in the background. (e) Our method successfully corrects the inaccurate depth values while in/outpainting each region with the guidance of the mask. Intermediate layered outputs are shown on the top right for the layered models.}
\label{fig:motivation}
\vspace{-0.4em}
\end{figure*}

Therefore, we take a novel approach of utilizing an additional cue of a \textit{high-quality mask} to refine depth maps predicted by SIDE methods. The provided mask can be hard (binary) or soft (e.g., from matting) and can be of objects or other parts of the image such as the sky. As high-quality auto-masking tools are very accessible nowadays, such masks can be easily obtained with commercial tools (e.g., \texttt{removebg} \cite{removebg} or Photoshop) or off-the-shelf segmentation models \cite{Yu_CVPR_2020, Yu_AAAI_2021, Zhuge_AAAI_2019, Hou_TPAMI_2019}. Segmentation masks can also be annotated by humans \cite{Xu_CVPR_2017, Cheng_TPAMI_2015, Wang_CVPR_2017}, and accurate datasets are easier to obtain than RGB-D data, which facilitates the training of auto-masking models.

However, even with such accurate masks, how to effectively train the depth refinement model remains an open issue. As shown in Figure~\ref{fig:motivation}(c), directly adding the mask as an input channel to the refinement model still results in blurrier boundaries than the given mask.
Therefore, we propose a \textit{layered refinement} strategy: The mask ($M$) and inverse mask ($1-M$) regions are processed separately to interpolate or extrapolate the depth values beyond the mask boundary, leading to two layers of depth maps. As shown in Figure~\ref{fig:motivation}(e), the refined output is the composite of the two layers using the mask $M$, which fully preserves the boundary details of the mask, as well as filling in the correct depth values for the isolated background regions.

A na\"ive baseline for layered depth refinement would be using an off-the-shelf inpainting method to generate the depth map layers for $M$ and $1-M$. Unfortunately, as shown in Figure~\ref{fig:motivation}(d), generic inpainting may not work well for filling in large holes in a depth map. Moreover, deriving an appropriate region for hole-filling on an imperfect initial depth prediction based on the mask is a non-trivial problem. The hole-filling region often needs to be expanded to cover uncertain regions along the mask boundary, as otherwise, the erroneous depth values may propagate in the hole. However, too much expansion will make the hole-filling task much more challenging as it may overwrite the original depth structure in the scene (see the $1-M$ layer in Figure~\ref{fig:motivation}(d)).

To address the challenge, 
we propose a framework for degradation-aware layered depth completion and refinement, which learns to identify and correct inaccurate regions based on the context of the mask and the image. Our framework does not require additional input or heuristics to expand the hole-filling region.
Furthermore, we devise a \textit{self-supervised} learning scheme that uses RGB-D training data without paired mask annotations.
We demonstrate that our method is robust under various conditions by empirically validating it on synthetic datasets and real images \textit{in the wild}. We further provide results on real-world downstream applications.

Our contributions are three-fold:
\begin{itemize}[noitemsep, topsep=0pt]
    \item We propose a novel mask-guided depth refinement framework that refines the depth estimations of SIDE models guided by a generic high-quality mask.
    \item We propose a novel \textit{layered refinement} approach, generating sharp and accurate results in challenging areas without additional input or heuristics. 
    \item We devise a self-supervised learning scheme that uses RGB-D training data \textit{without} paired mask annotations.
\end{itemize}

\section{Related Work}
\noindent
\textbf{Single Image Depth Estimation}\quad
Single image depth estimation (SIDE), also commonly termed monocular depth estimation, aims to predict a depth map from an RGB image. A common approach is to train a deep neural network on RGB-D datasets to learn the non-linear mapping from RGB to depth \cite{Li_CVPR_2018, Xian_2020_CVPR, Ranftl_TPAMI_2020, Ranftl_ICCV_2021}. As for the model architecture, convolutional neural networks (CNNs) are a popular choice \cite{Ranftl_TPAMI_2020, Xian_2020_CVPR}, with a transformer-based model \cite{Ranftl_ICCV_2021} being recently proposed to overcome the limited receptive field size of CNNs. Transformer models \cite{Dosovitskiy_ICLR_2021, Tolstikhin_arxiv_2021} leverage self-attention \cite{Vaswani_NIPS_2017}, expanding the receptive field to the entire image at every level. We also base our model architecture on transformers to benefit from the enlarged receptive field. 

For training SIDE models, datasets are often augmented with synthetic datasets \cite{Butler_ECCV_2012, Wang_ARXIV_2019, Wang_IROS_2020, Yao_CVPR_2020, Niklaus_TOG_2019} and relative depths computed from stereo images \cite{Li_CVPR_2018, Xian_CVPR_2018, Wang_3DV_2019}. Numerous supervision schemes \cite{Garg_ECCV_2016, Mousavian_OTHER_2016, Godard_CVPR_2017, Zhou_CVPR_2017, Abarghouei_CVPR_2018, Mahjourian_CVPR_2018, Zheng_ECCV_2018, Wong_CVPR_2019, Zhu_CVPR_2020, Chen_ECCV_2020} and loss functions \cite{Li_CVPR_2018, Jiao_ECCV_2018, Lee_ECCV_2020, Xian_2020_CVPR} have been proposed to optimize the model training for SIDE. Several methods \cite{Mousavian_OTHER_2016, Zhu_CVPR_2020, Wang_CVPR_2020} attempt to exploit the relation between image segmentation and SIDE, with Zhu \textit{et al.} \cite{Zhu_CVPR_2020} proposing regularizing depth boundaries with segmentation map boundaries in the loss function to enforce sharper edges in the resulting depth maps. However, even with sophisticated framework designs, capturing highly accurate depth boundaries remains a challenge due to the ill-posed nature of the problem and the lack of pixel-perfect ground truth depth data.

\smallskip\noindent
\textbf{Depth Inpainting}\quad
Inpainting depth maps is often necessary in novel view synthesis for 3D photography to naturally fill in disoccluded regions \cite{Niklaus_TOG_2019, Shih_CVPR_2020, Jampani_ICCV_2021}. Such methods apply joint RGB and depth inpainting in the background region near object edges. Another line of research is depth completion, which aims to fill in unknown depth values from sparsely known annotations. Imran~\textit{et al.}~\cite{Imran_CVPR_2021} proposed a layered approach, extrapolating foreground and background regions separately from LiDAR data. In our depth refinement method, both the mask and inverse mask regions are inpainted/outpainted \textit{while} correcting inaccurate depth values and merged afterward to obtain accurate boundaries.

\smallskip\noindent
\textbf{Depth Refinement}\quad
In an inspirational work \cite{Miangoleh_CVPR_2021}, Miangoleh \textit{et al.} proposed boosting high-frequency details in SIDE results by merging multiple depth predictions at various resolutions, exploiting the limited receptive field size of CNNs. However, their merging algorithm tends to generate inconsistent depth values in foreground objects, and its refinement degrades with recent transformer architectures as it is based on a fundamental assumption related to CNNs. Furthermore, capturing very thin boundaries and generating accurate depth values in hole regions are still challenging. 

In this paper, we explore a novel direction of using generic masks as guidance for depth refinement. Unlike previous methods that upscale or enhance details in the entire depth map, we focus on delicately refining along the boundary and hole regions of the mask. Handling such regions is often important in downstream applications such as Bokeh effect synthesis. Our method is generic and can refine depth maps generated by any SIDE model regardless of the model architecture, as long as the provided mask contains better boundaries than the initial depth map. Note that our method operates in the inverse depth space as many prior works \cite{Miangoleh_CVPR_2021, Ranftl_TPAMI_2020, Ranftl_ICCV_2021}, although we continue using the term depth.

\section{Proposed Method}
We propose a layered depth refinement framework for enhancing the initial depth prediction of SIDE models using the guidance of a quasi-accurate mask and an RGB image.

\begin{figure}
\centering
\includegraphics[width=0.99\linewidth]{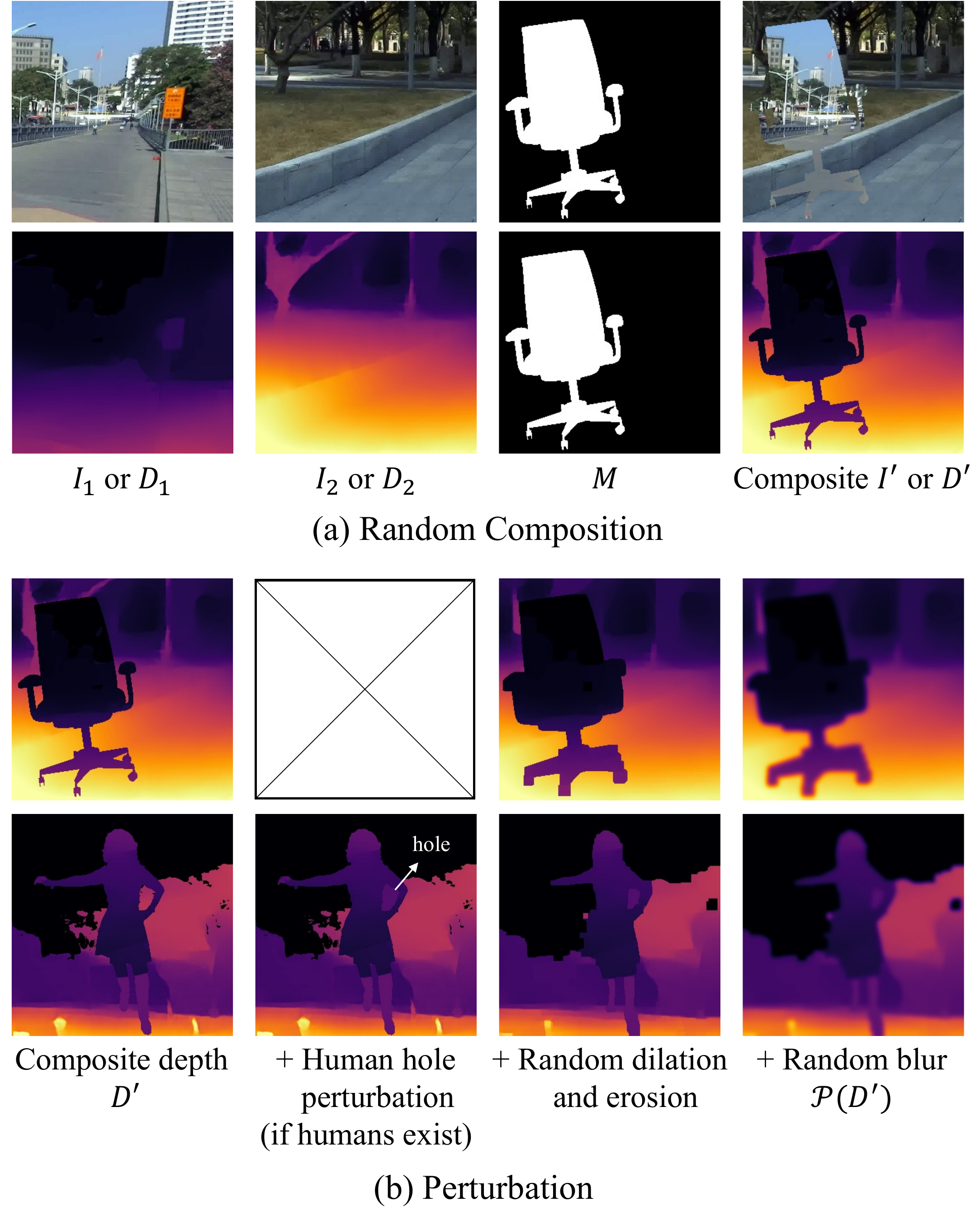}
\caption{Data generation scheme. RGB-D patches are randomly composited using an arbitrary binary mask. Perturbations are applied to simulate depth estimates, resulting in isolated regions being covered up and thin structures being lost.}
\label{fig:data_gen}
\end{figure}

\subsection{Data Generation}
\noindent
\textbf{Random composition}\quad
With an RGB-D dataset consisting of an RGB image $I$ and its depth map $D$, a general depth refinement model can be optimized in a self-supervised way by applying random perturbations $\mathcal{P}$ on $D$, which inversely simulate initial depth predictions. A neural network $\mathcal{R}$ can then be trained to predict the refined depth map $\hat{D}=\mathcal{R}(\mathcal{P}(D), I)$ with an appropriate loss function $\mathcal{L}(\hat{D}, D)$. 

However, collecting a dataset for training a \textit{mask-guided} depth refinement model is challenging as datasets containing masks along with the RGB-D information are scarce. Hence, we devise a data generation scheme that does \textit{not} require paired depth and mask annotations. Specifically, a composite depth map $D'$ is randomly synthesized from two arbitrary depth maps $D_1$ and $D_2$ using an arbitrary binary mask $M$ with $m_{ij}\in\{0, 1\}$, by $D' = M\cdot D_1 + (1-M)\cdot D_2$. Likewise, the corresponding composite RGB image $I'$ is computed by $I' = M\cdot I_1 + (1-M)\cdot I_2$,
where $I_1$ and $I_2$ are the RGB images corresponding to $D_1$ and $D_2$, respectively. Examples of $D'$ and $I'$ are shown in Figure~\ref{fig:data_gen}(a).
Applying perturbations to $D'$ leads to $\mathcal{P}(D')$, and the mask-guided refinement model $\mathcal{R}_m$ can then be trained with $\mathcal{L}(\hat{D}', D')$, where $\hat{D}'=\mathcal{R}_m(\mathcal{P}(D'), I', M)$. 

\begin{figure*}
\centering
\includegraphics[width=\linewidth]{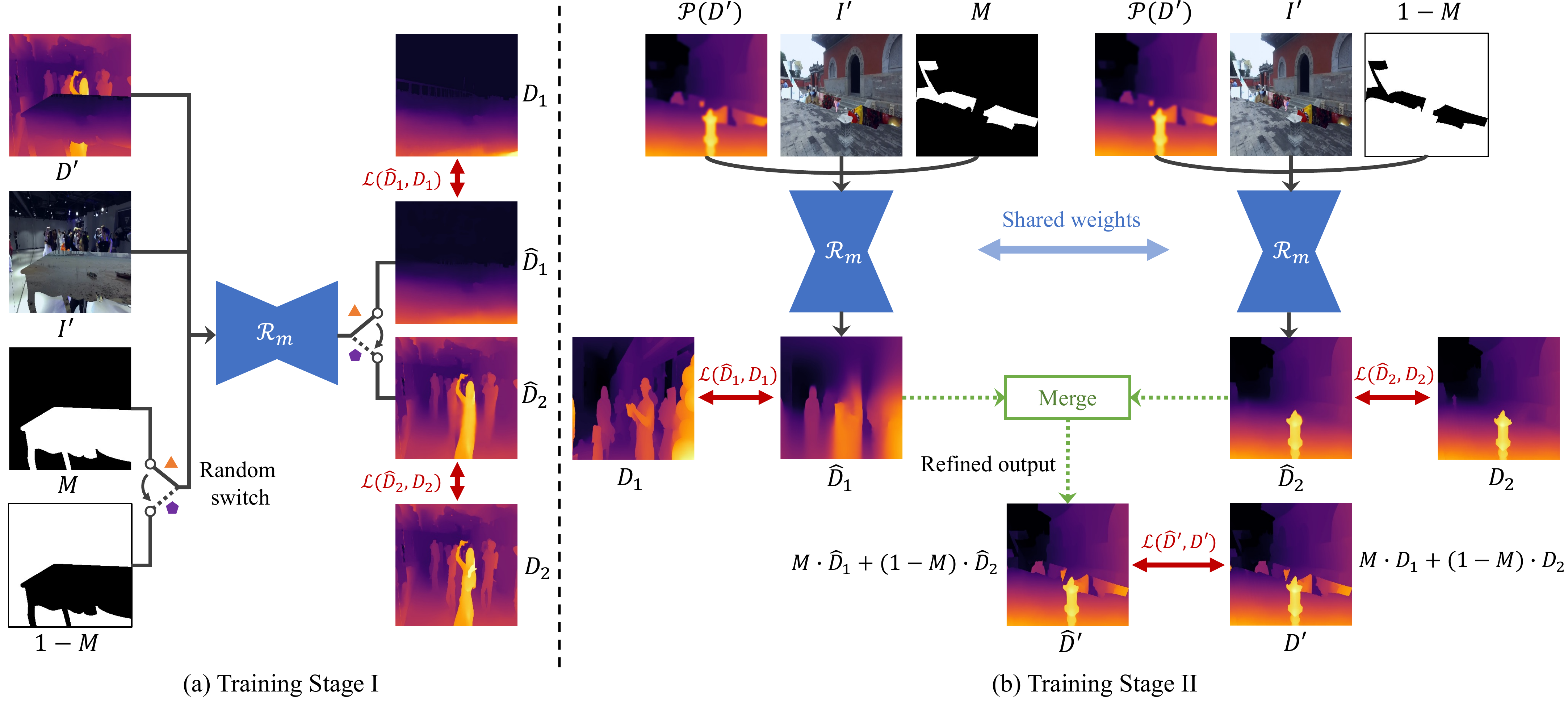}
\caption{An overview of the proposed two-stage training strategy. In the first stage, the network is trained to complete regions with 0 based on regions with 1 in the given mask. The mask is randomly flipped and the corresponding depth ($D_1$ or $D_2$) is given as the ground truth. In stage II, we run the network twice to obtain $\hat{D}_1$ and $\hat{D}_2$ and merge them based on the mask to produce the refined output $\hat{D}'$. The network learns to remove perturbations while inpainting/outpainting the depth. During inference, the refined output is obtained following stage II.}
\label{fig:train_strat}
\vspace{-0.1em}
\end{figure*}

In this way, we can obtain a synthesized depth map $D'$ and an RGB image $I'$ that are aligned to $M$ from any RGB-D dataset and arbitrary masks. Diverse types of masks can be mixed and used, including object and stuff masks from segmentation datasets \cite{Lin_ARXIV_2014, Zhoub_CVPR_2017}. Furthermore, we can effortlessly acquire the ground truths for inpainting/outpainting ($D_1$ and $D_2$), which are essential for our layered refinement approach, explained in more detail in the next section.

\smallskip\noindent
\textbf{Perturbations}\quad
As shown in Figure~\ref{fig:data_gen}(b), we apply three types of perturbations on $D'$ to simulate typical inaccuracies in SIDE model predictions. First, random dilation and erosion are applied in a random order so that the perturbed depth map lacks thin structures, and its depth boundaries are not always aligned with the RGB image or the mask. In Figure~\ref{fig:data_gen}(b), it can be observed that thin structures (hand of the person) are lost, and isolated regions are covered up (between the arm and the main frame of the chair) after random dilation and erosion. Second, we apply random amounts of Gaussian blur on the depth map as estimated depth maps tend to have blurry boundaries. Lastly, we design a \textit{human hole perturbation scheme} that detects isolated regions and assigns a random value between the mean depth values surrounding the hole and inside the original hole, simulating the often-missing isolated regions inside human bodies in estimated depth maps. 
More details of the perturbation scheme are provided in the appendix.

\subsection{Training Strategy} \label{sec:train_strat}
\noindent
\textbf{Two-stage training for layered refinement}\quad
Although depth refinement with an accurate mask may appear straightforward after data pairs are obtained with the proposed data generation scheme, directly predicting the refined depth map from concatenated RGB-D and mask inputs leads to suboptimal results, as shown in Figure~\ref{fig:motivation}. To explicitly benefit from the accurate mask, we propose a \textit{layered refinement} approach that refines regions specified by $M$ and $1-M$ \textit{separately} and merges two individual results based on $M$. In this way, the model can focus on correcting the depth values in each region, and mask boundaries can be fully preserved after the merging stage.

We train our model in two stages shown in Figure~\ref{fig:train_strat}. In the first stage, the model $\mathcal{R}_m$ is trained for image completion by randomly providing $M$ or $1-M$ and optimizing either $\mathcal{L}(\mathcal{R}_m(D', I', M), D_1)$ or $\mathcal{L}(\mathcal{R}_m(D', I', 1-M), D_2)$. Note that a single model is trained for both inpainting \textit{and} outpainting the depth input to always complete regions with $0$ based on regions with $1$ signified by the given mask $M$ or $1-M$. 
Then in the second stage, we add perturbations $\mathcal{P}$ and run the network twice with $M$ and $1-M$ to obtain two outputs $\hat{D}_1$ and $\hat{D}_2$, given by
\begin{align} \label{eq:layered}
    \hat{D}_1 &= \mathcal{R}_m(\mathcal{P}(D'), I', M) \qquad \text{and} \\
    \hat{D}_2 &= \mathcal{R}_m(\mathcal{P}(D'), I', 1-M).
\end{align}
Reasonable $\hat{D}_1$ and $\hat{D}_2$ are generated from the beginning of the second stage as the model has been pretrained for inpainting/outpainting in the first stage. Finally, $\hat{D}_1$ and $\hat{D}_2$ are merged to yield the refined output $\hat{D}'$ as follows:
\begin{equation} \label{eq:merge}
    \hat{D}' = M\cdot \hat{D}_1 + (1-M)\cdot \hat{D}_2.
\end{equation}

Our model is optimized with three losses at this stage: $\mathcal{L}(\hat{D}_1, D_1)$, $\mathcal{L}(\hat{D}_2, D_2)$, and $\mathcal{L}(\hat{D}', D')$. As a result, the network learns to remove perturbations while generating completed depth maps under a \textit{unified} framework. Although we only utilize composite depth maps as input during training, the randomness in composition (\textit{random} depth maps composited with a \textit{random} mask) and random perturbations lead to a robust model that generalizes well to real depth estimations and diverse masks.

\begin{figure}
\centering
\includegraphics[width=0.94\linewidth]{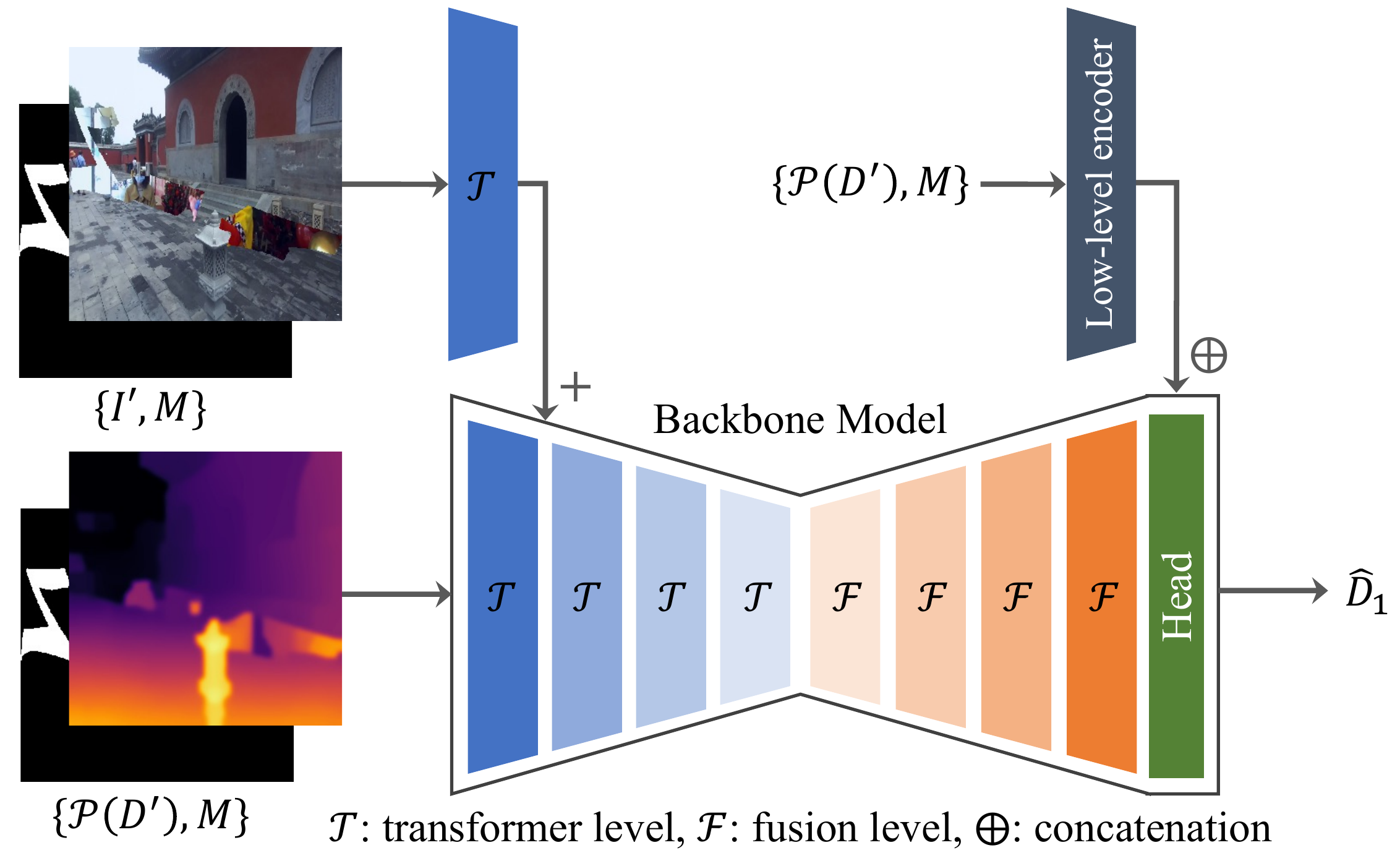}
\caption{Our network architecture with DPT \cite{Ranftl_ICCV_2021} as the backbone model. We add a low-level encoder and a branch for the RGB input.
}
\label{fig:net_arch}
\end{figure}

\smallskip\noindent
\textbf{Loss function}\quad
The loss $\mathcal{L}$ is comprised of three different loss terms summed with unit scale: L1 loss, L2 loss, and a multi-scale gradient loss with four scale levels \cite{Li_CVPR_2018}. The gradient loss is adopted to enforce sharp depth boundaries.

\subsection{Model Architecture}
We base our model architecture on the dense prediction transformer (DPT) \cite{Ranftl_ICCV_2021} with four transformer encoder levels \cite{Dosovitskiy_ICLR_2021} $l\in\{1, 2, 3, 4\}$ and four fusion decoder levels. At each encoder level, overlapping patches are extracted and embedded to dimensions $d_l\in\{64, 128, 320, 512\}$ and fed into $t_l\in\{3, 4, 18, 3\}$ transformer layers each with self-attention, LayerNorm \cite{Ba_ARXIV_2016} and MLP layers. The spatial resolution is decreased by a scale factor of $s_l\in\{4, 2, 2, 2\}$ at each level. On the decoder side, features are fused with residual convolutional units at each fusion level, followed by a monocular depth estimation head at the end as in \cite{Ranftl_ICCV_2021}.

As shown in Figure~\ref{fig:net_arch}, we insert an additional encoder branch with a single transformer level to the original backbone so that $D'$ (or $\mathcal{P}(D')$) and $M$ (or $1-M$) are concatenated and fed into the main branch, and $I'$ concatenated with $M$ (or $1-M$) are fed into the additional branch. The outputs are simply summed after the first transformer level. Additionally, a lightweight low-level encoder is introduced to encode the low-level features of the input depth map. These features are concatenated with the features from the main decoder branch and entered into the head, ensuring that the network does not forget the initial depth values.

\section{Experiments}
\subsection{Implementation Details}
We train our model for 500K iterations for the first stage and another 500K iterations for the second stage following the training strategy described in Sec. \ref{sec:train_strat}. We used a training patch size of $320\times 320$ and a batch size of 32. The model is optimized with AdamW \cite{Loshchilov_ICLR_2019} at an initial learning rate of $10^{-4}$, which is decreased by 1/10 at 60\% and 80\% of the total number of iterations. Our model is implemented using PyTorch and trained on 4 NVIDIA V100 GPUs.

For data augmentation, we apply random horizontal flipping and resizing to the input depth maps and RGB images. RGB images are further augmented with random contrast, saturation, brightness, JPEG compression, and grayscale conversions to make our model more robust to various types of inputs. Our model is trained on diverse indoor and outdoor natural RGB-D images, with depth maps scaled to $[0, 10]$ as in \cite{Yin_CVPR_2021} and RGB images normalized using ImageNet \cite{Deng_CVPR_2009} mean and standard deviation. Furthermore, to benefit from the proposed self-supervised learning scheme that supports diversifying the types of masks, we sample 50\% of masks from diverse object masks, 20\% from sky masks and 30\% from human masks, where humans with holes are selected 50\% of the time (15\% of all masks) during training.

\begin{table*}\centering
    \ra{1.1}
    \scalebox{0.8}{
    \begin{tabular}{lggggcccggggcc}
    \toprule
    \multirow{2}{*}{Method} & \multicolumn{6}{c}{Hypersim \cite{Roberts_ICCV_2021}} & \phantom{a} & \multicolumn{6}{c}{TartanAir \cite{Wang_IROS_2020}} \\
    \cmidrule{2-7} \cmidrule{9-14}
    & $\text{R}^3\uparrow$ & MBE$\downarrow$ & $\varepsilon_{acc}\downarrow$ & $\varepsilon_{comp}\downarrow$ & WHDR$\downarrow$ & RMSE$\downarrow$ && $\text{R}^3\uparrow$ & MBE$\downarrow$ & $\varepsilon_{acc}\downarrow$ & $\varepsilon_{comp}\downarrow$ & WHDR$\downarrow$ & RMSE$\downarrow$ \\
    \midrule
    MiDaS v2.1 \cite{Ranftl_TPAMI_2020} & - & 0.0973 & 2.521 & 7.074 & 0.1496 & 0.0966 && - & 0.0596 & 3.483 & 6.913 & \textbf{0.1207} & 0.0533 \\
    \midrule
    + Direct-\textit{composite} & 3.771 & 0.0941 & 1.915 & 6.233 & 0.1490 & 0.0961 && 5.897 & 0.0594 & 3.183 & 6.363 & 0.1209 & 0.0534 \\
    + Direct-\textit{paired} & - & - & - & - & - & - && 3.507 & 0.0575 & 3.153 & 6.304 & 0.1196 & \textbf{0.0525} \\
    + Layered-\textit{propagation} & 1.097 & 0.1044 & 1.942 & 6.284 & 0.1629 & 0.1028 && 3.642 & 0.0608 & 3.128 & 6.358 & 0.1255 & 0.0550 \\
    + Layered-\textit{ours} & 2.332 & 0.1000 & \textbf{1.871} & 6.396 & 0.1560 & 0.0999 && 6.939 & 0.0580 & 3.243 & 6.437 & 0.1230 & 0.0539 \\
    + Ours (proposed) & \textbf{5.209} & \textbf{0.0906} & 1.888 & \textbf{5.931} & \textbf{0.1481} & \textbf{0.0958} && \textbf{16.569} & \textbf{0.0579} & \textbf{2.851} & \textbf{6.272} & \textbf{0.1207} & 0.0538 \\
    \midrule \midrule
    DPT-Large \cite{Ranftl_ICCV_2021} & - & 0.0936 & 2.071 & 6.190 & 0.1347 & 0.0911 && - & 0.0496 & 2.574 & 5.677 & 0.1091 & 0.0414 \\
    \midrule
    + Direct-\textit{composite} & 2.574 & 0.0891 & 1.599 & 5.411 & 0.1339 & 0.0903 && 4.773 & 0.0486 & 2.462 & 5.480 & 0.1086 & 0.0411 \\
    + Direct-\textit{paired} & - & - & - & - & - & - && 2.413 & 0.0485 & 2.519 & 5.394 & 0.1105 & 0.0412 \\
    + Layered-\textit{propagation} & 1.188 & 0.1007 & 1.792 & 5.636 & 0.1502 & 0.0986 && 2.347 & 0.0524 & 2.579 & 5.527 & 0.1162 & 0.0442 \\
    + Layered-\textit{ours} & 1.996 & 0.0954 & 1.606 & 5.605 & 0.1433 & 0.0953 && 5.626 & 0.0484 & 2.447 & 5.342 & 0.1116 & 0.0423 \\
    + Ours (proposed) & \textbf{4.455} & \textbf{0.0840} & \textbf{1.491} & \textbf{5.087} & \textbf{0.1333} & \textbf{0.0896} && \textbf{8.767} & \textbf{0.0474} & \textbf{2.282} & \textbf{5.245} & \textbf{0.1078} & \textbf{0.0408} \\
    \bottomrule
    \end{tabular}}
    \caption{Quantitative results on Hypersim \cite{Roberts_ICCV_2021} and TartanAir \cite{Wang_IROS_2020} comparing mask-guided depth refinement models. Best values in \textbf{bold}.} \label{table:quant_eval}
\end{table*}

\subsection{Evaluation Datasets} 
For a quantitative evaluation, datasets with both depth and mask annotations are needed to exclude potential errors caused by inaccurate masking. Furthermore, the ground truth depth should be accurate for reliable evaluations on fine boundaries and object holes. Thus, we use Hypersim (CC-BY SA 3.0 License) \cite{Roberts_ICCV_2021} and TartanAir (3-Clause BSD License) \cite{Wang_IROS_2020}, which are recently released synthetic datasets that contain dense and accurate depth values and also have instance segmentation maps. We select the first frame in each camera trajectory per scene for Hypersim and the 100-th frame for each trajectory in $Easy$ difficulty per environment for TartanAir as the test set, which results in 456 images and 206 images in total for Hypersim and TartanAir, respectively. Other datasets such as Cityscapes \cite{Cordts_CVPR_2016} are not appropriate as the ground truth depth is noisy, often inaccurate around edges and misses thin structures. Additionally, we qualitatively evaluate our refinement method on various freely licensed images from the web \cite{unsplash, pixabay} with an auto-masking tool \cite{removebg}.

\smallskip\noindent
\textbf{Zero-shot cross-dataset transfer}\quad
We follow the experiment protocol in \cite{Ranftl_TPAMI_2020} for evaluation. None of the compared methods or our method have seen the RGB-D images in Hypersim \cite{Roberts_ICCV_2021} or TartanAir \cite{Wang_IROS_2020} during training. Predictions are scaled and shifted using $l2$ minimization to match the ground truth depth. 

\smallskip\noindent
\textbf{Inference using segmentation maps}\quad
To use segmentation maps in a mask-guided framework, we take the following steps: (i) compute a binary mask $M_i$ for each instance $i$ with more than 1\% of the total number of pixels in the instance segmentation map, (ii) run the model $N$ times with $M_i$, and (iii) merge the refined outputs $\hat{D}_i$ per each pixel by $\hat{D}=\argmax_{\hat{D}_i}(|D'-\hat{D}_i|)$, where $D'$ is initial depth.

\subsection{Evaluation Metrics}
We evaluate the overall error of the output depth maps with the RMSE and the Weighted Human Disagreement Rate (WHDR) \cite{Chen_NIPS_2016} measured on 10K randomly sampled point pairs. To evaluate the boundary quality, we report the depth boundary error \cite{Koch_ECCVW_2018} on accuracy ($\varepsilon_{acc}$) and completeness ($\varepsilon_{comp}$). In addition, we propose two metrics, mask boundary error (MBE) and relative refinement ratio ($\text{R}^3$). All metrics are measured in the inverse depth space.

MBE computes the average RMSE on mask boundary pixels over the $N$ instances. Mask boundary $M_i^b$ is obtained by subtracting the eroded $M_i$ from $M_i$ and dilating it with a $5\times 5$ kernel. The MBE is then given by
\begin{align} \label{eq:mbe}
    \text{MBE} &= \frac{1}{N}\sideset{}{_{i=1}^{N}}\sum\sqrt{\frac{1}{N_i^b}\sum(M_i^b\cdot D-M_i^b\cdot\hat{D})^2},
\end{align}
where $N_i^b$ is the number of boundary pixels for each instance $i$. With $\varepsilon_{acc}$, $\varepsilon_{comp}$ and MBE, we can comprehensively measure the boundary accuracy of the refined depth map: $\varepsilon_{acc}$ and $\varepsilon_{comp}$ focusing on \textit{depth} boundaries and MBE on the \textit{mask} boundaries of depth maps. Furthermore, we define $\text{R}^3$ (relative refinement ratio) as the ratio of the number of pixels improved by more than a threshold $t$ to the number of pixels worsened by more than $t$, in terms of absolute error.
We set $t=0.05$ and compute $\text{R}^3$ of refined results over initial results by base models \cite{Ranftl_TPAMI_2020, Ranftl_ICCV_2021}. $\text{R}^3$ is a meaningful indicator for assessing the refinement performance.

\begin{figure*}
\centering
\includegraphics[width=\linewidth]{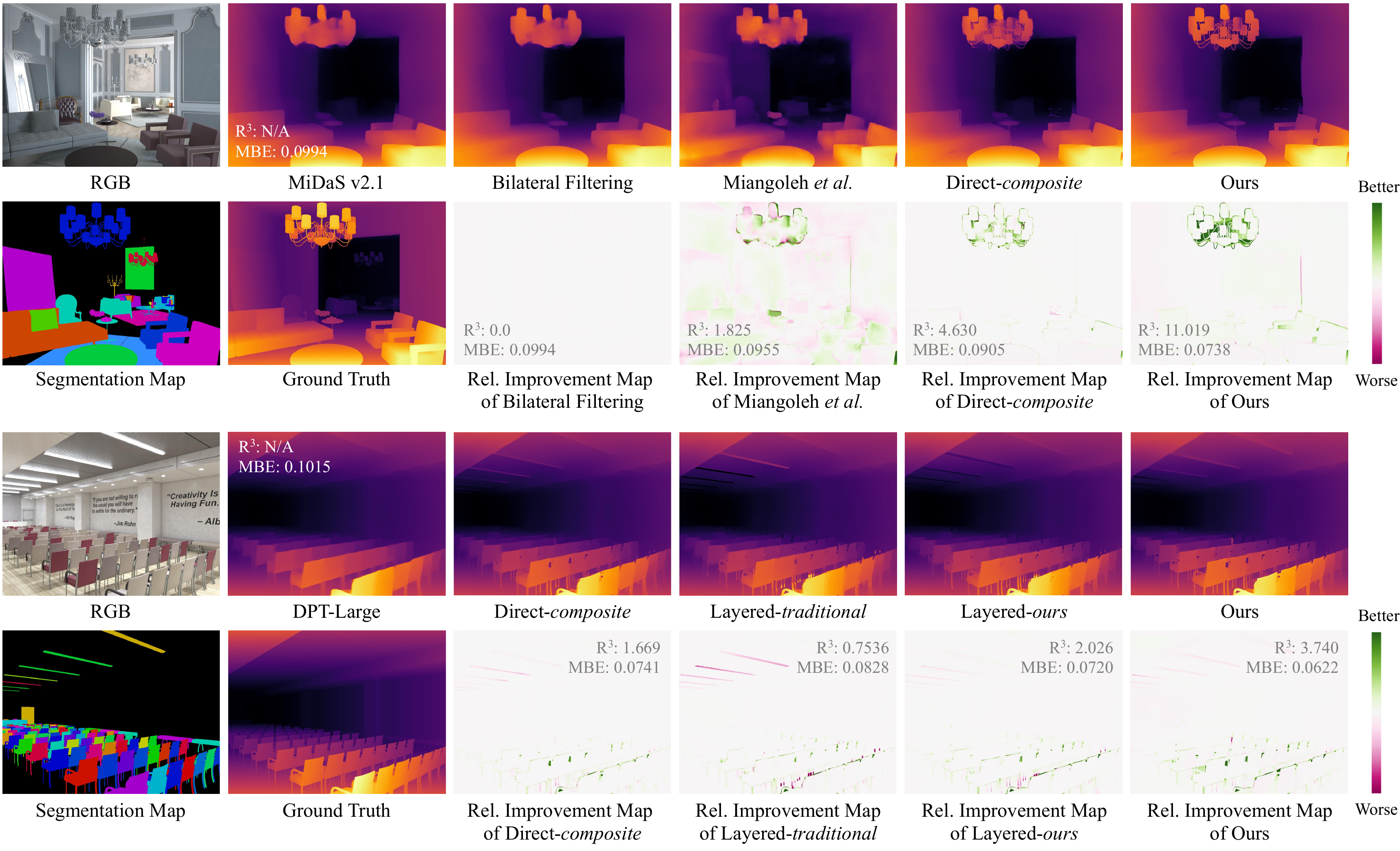}
\caption{Qualitative results on Hypersim \cite{Roberts_ICCV_2021}. The relative improvement maps visualize where the refinement method improved and worsened the initial depth estimation by \cite{Ranftl_TPAMI_2020} or \cite{Ranftl_ICCV_2021}. Our method focuses on the edges and hole regions, accurately refining fine structures.}
\label{fig:qual_hypersim}
\end{figure*}

\subsection{Compared Methods} \label{sec:comp_methods} 
To evaluate the refinement performance, we apply our method to the initial depth predictions of two SIDE models: CNN-based MiDaS v2.1 \cite{Ranftl_TPAMI_2020} and SOTA transformer-based DPT-Large \cite{Ranftl_ICCV_2021}. Since there are no existing methods that perform mask-guided depth refinement, we set up the following baselines using masks for comparison:
\begin{itemize}[noitemsep] 
    \item Direct-\textit{composite} produces the refined output without layering and is trained on the same dataset as ours (with composite images and the mask).
    \item Direct-\textit{paired} also refines without layering but is trained on paired RGB-D and masks in Hypersim \cite{Roberts_ICCV_2021}. Hence, we only evaluate on TartanAir \cite{Wang_IROS_2020} for this method.
    \item Layered models (Layered-\textit{propagation} and Layered-\textit{ours}) either apply a propagation-based image completion algorithm \cite{Telea_JGT_2004} or use our model from stage I training, once with the dilated mask for inpainting and the second time with the eroded mask for outpainting. The inpainted/outpainted results are then merged with the mask, similar to our proposed approach.
\end{itemize}
The network architecture used for Direct-\textit{composite} and Direct-\textit{paired} is the same as our encoder-decoder-style transformer model in Figure~\ref{fig:net_arch}. For the layered models, we set the dilation and erosion kernel to $5\times 5$ for evaluation with segmentation maps. For images \textit{in the wild}, we manually tweak the kernel sizes for each image to obtain the best results.

Additionally, we compare to bilateral median filtering (BMF) with parameters from \cite{Shih_CVPR_2020} (previously used for refining depth maps in \cite{Ma_ICCV_2013, Shih_CVPR_2020}) and Miangoleh \textit{et al.}'s recent depth refinement method \cite{Miangoleh_CVPR_2021}. These approaches do not use masks as guidance.
For all compared methods, we use the officially released code and weights. 

\subsection{Analysis}

In Table \ref{table:quant_eval}, we provide the quantitative results on mask-guided refinement methods. Our method improves both MiDaS v2.1 \cite{Ranftl_TPAMI_2020} and DPT-Large \cite{Ranftl_ICCV_2021} on all edge-related metrics ($\varepsilon_{acc}$, $\varepsilon_{comp}$ and MBE) and results in high $\text{R}^3$ values of at most $16.569$.
WHDR and RMSE values are not very discriminative between mask-guided refinement methods as they measure the average error over all pixels, whereas mask-guided refinement methods aim at refining along mask boundaries and leave most internal regions as is. Our method outperforms all baselines in $\text{R}^3$ and MBE, demonstrating the power of our layered refinement approach.

In Table \ref{table:quant_eval_nomask}, we compare to automatic depth refinement methods without mask-guidance. Conventional image filtering fails to enhance the edge-related metrics.
Miangoleh \textit{et al.}'s method \cite{Miangoleh_CVPR_2021} is at times better on the global edge metrics ($\varepsilon_{acc}$ and $\varepsilon_{comp}$) as it enhances all edges in the depth map. However, as it also carries the risk of distorting the original values, $\text{R}^3$ values tend to be lower compared to ours, which mostly refines along mask boundaries and leaves other regions intact.
Furthermore, as \cite{Miangoleh_CVPR_2021} heavily relies on the base model's behavior, its generalization capability is limited for other architecture types such as a transformer \cite{Ranftl_ICCV_2021}. Our method works well regardless of the base model architecture and generalizes well to both datasets, leading to the best metric values when coupled with \cite{Ranftl_ICCV_2021}. 

\begin{table}\centering
    \scalebox{0.66}{
    \begin{tabular}{lcccccccc}
    \toprule
     & \multicolumn{4}{c}{Hypersim \cite{Roberts_ICCV_2021}} & \multicolumn{4}{c}{TartanAir \cite{Wang_IROS_2020}} \\
    & $\text{R}^3\uparrow$ & MBE$\downarrow$ & $\varepsilon_{acc}\downarrow$ & $\varepsilon_{comp}\downarrow$ & $\text{R}^3\uparrow$ & MBE$\downarrow$ & $\varepsilon_{acc}\downarrow$ & $\varepsilon_{comp}\downarrow$ \\
    \midrule
    \cite{Ranftl_TPAMI_2020} & - & 0.0973 & 2.521 & 7.074 & - & 0.0596 & 3.483 & 6.913 \\
    \midrule 
    + BMF & 0.7784 & 0.0974 & 2.574 & 7.089 & 1.032 & 0.0597 & 3.489 & 6.947 \\
    + \cite{Miangoleh_CVPR_2021} & 4.671 & 0.0923 & \textbf{1.551} & \textbf{5.837} & 4.721 & 0.0602 & 3.605 & 7.287 \\
    + Ours & \textbf{5.209} & \textbf{0.0906} & 1.888 & 5.931 & \textbf{16.569} & \textbf{0.0579} & \textbf{2.851} & \textbf{6.272} \\
    \midrule\midrule
    \cite{Ranftl_ICCV_2021} & - & 0.0936 & 2.071 & 6.190 & - & 0.0496 & 2.574 & 5.677 \\
    \midrule
    + BMF & 0.9444 & 0.0937 & 2.094 &  6.203 & 0.6875 & 0.0497 & 2.667 & 5.836 \\
    + \cite{Miangoleh_CVPR_2021} & 1.843 & 0.0905 & 1.681 & 5.633 & 4.013 & 0.0496 & 2.414 & 5.569 \\
    + Ours & \textbf{4.455} & \textbf{0.0840} & \textbf{1.491} & \textbf{5.087} & \textbf{8.767} & \textbf{0.0474} & \textbf{2.282} & \textbf{5.245} \\
    \bottomrule
    \multicolumn{9}{l}{BMF: Bilateral Median Filtering} \\
    \end{tabular}}
    \caption{Comparison to automatic refinement methods. Our method refines mask boundaries and leaves other regions intact whereas \cite{Miangoleh_CVPR_2021} refines all regions at the risk of distorting original values.
    } 
    \label{table:quant_eval_nomask}
\end{table}

In Figure~\ref{fig:qual_hypersim}, we show the qualitative results on Hypersim \cite{Roberts_ICCV_2021}. We also visualize the relative improvement maps showing where the absolute error decreased compared to the base model MiDaS v2.1 \cite{Ranftl_TPAMI_2020} or DPT \cite{Ranftl_ICCV_2021}. Our method focuses on refining edges and hole regions and leaves most other regions untouched, whereas Miangoleh \textit{et al.}'s method \cite{Miangoleh_CVPR_2021} often worsens homogeneous regions. Compared to other baselines, our layered refinement approach within a unified framework helps to correct low-level details effectively.

\begin{figure}
\centering
\includegraphics[width=\linewidth]{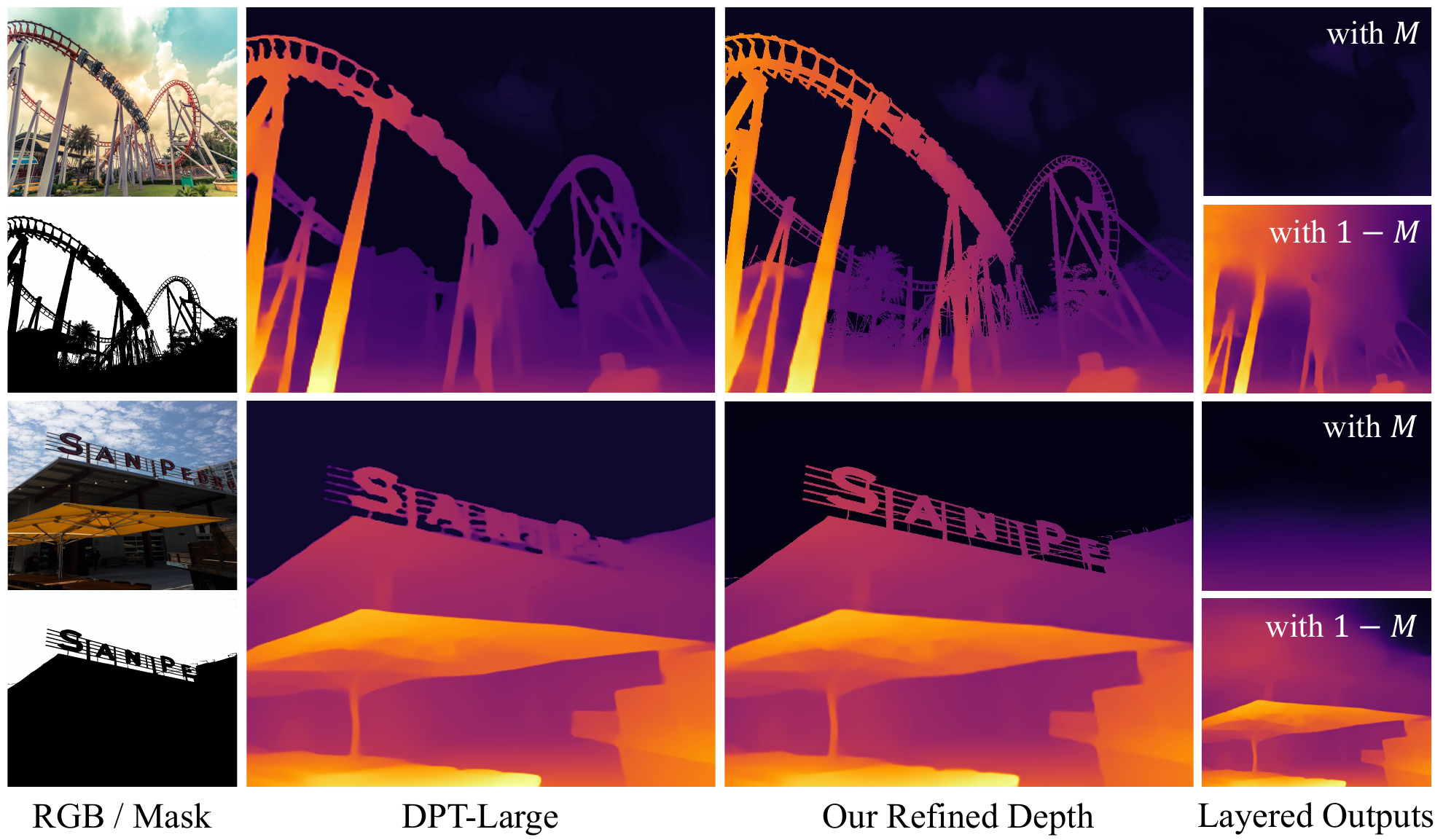}
\caption{Refined results on real images with various masks.}
\label{fig:qual_sky}
\end{figure}

\begin{table}\centering
    \scalebox{0.8}{
    \begin{tabular}{ccc|cccc}
    \toprule
    Stage I & Stage II & HP & $\text{R}^3\uparrow$ & MBE$\downarrow$ & $\varepsilon_{acc}\downarrow$ & $\varepsilon_{comp}\downarrow$ \\
    \midrule
    \rowcolor[gray]{.95}\multicolumn{3}{c}{DPT-Large \cite{Ranftl_ICCV_2021}} & - & 0.0936 & 2.071 & 6.190 \\
    \cmark &  &  & 1.996 & 0.0954 & 1.606 & 5.605 \\
    & \cmark &  & 2.016 & 0.0890 & 1.915 & 5.320 \\
    & \cmark & \cmark & 2.613 & 0.0861 & 1.670 & 5.161 \\
    \cmark & \cmark & & \textbf{5.384} & 0.0846 & \textbf{1.438} & 5.100 \\
    \cmark & \cmark & \cmark & 4.455 & \textbf{0.0840} & 1.491 & \textbf{5.087} \\
    \bottomrule
    \multicolumn{7}{l}{HP: Hole Perturbation} \\
    \end{tabular}}
    \caption{Ablation study on Hypersim \cite{Roberts_ICCV_2021}. Best values in \textbf{bold}.} \label{table:ablation2}
\end{table}

\begin{figure}
\centering
\includegraphics[width=\linewidth]{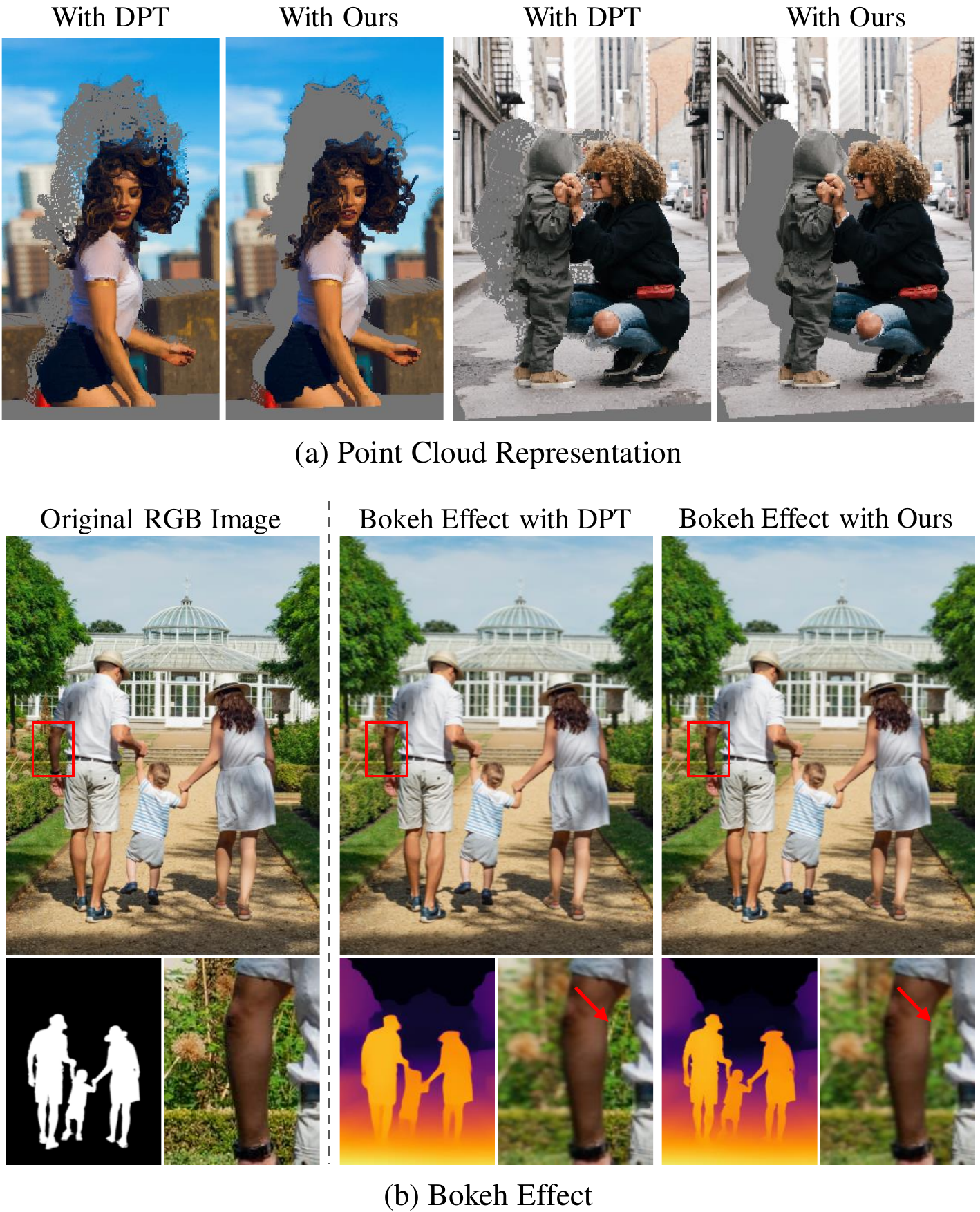}
\caption{Point cloud and Bokeh effect \cite{Xiao_TOG_2018} using initial depth by DPT \cite{Ranftl_ICCV_2021} and refined depth by Ours. Better viewed with zoom-in. 
}
\label{fig:applications}
\end{figure}

\smallskip\noindent
\textbf{Images in the wild}\quad
We further evaluate our model on real images \textit{in the wild} to assess its generalization ability and robustness. Comparisons to baselines are shown in Figure~\ref{fig:motivation} and more results are shown in Figures~\ref{fig:teaser}, \ref{fig:qual_sky}, and \ref{fig:applications}. Our method is able to generate sharp depth maps consistent with the mask for various real images. All portrait images are free-licensed images from \texttt{unsplash} \cite{unsplash} and \texttt{pixabay} \cite{pixabay}, and masks are generated with \texttt{removebg} \cite{removebg}. Sky images are licensed by Adobe Stock \cite{AdobeStock}, and their masks are annotated using a commercial photo editing tool. 

\smallskip\noindent
\textbf{Ablation study}\quad
We provide an ablation study of our model in Table \ref{table:ablation2} by removing different components in our framework. Stage I helps start from better-initialized parameters, and Stage II is necessary to train our model for layered refinement under a unified framework. Ablating either of them results in performance degradation. Although the quantitative results with or without hole perturbations are similar, hole perturbations are crucial in improving holes in humans.

\smallskip\noindent
\textbf{Results on downstream applications}\quad
More accurate depth maps can improve the outcomes of downstream applications. In Figure~\ref{fig:applications}(a), edges and holes are improved with our refined depth map in point cloud representations of a novel view. In Figure~\ref{fig:applications}(b), we apply Bokeh effect \cite{Xiao_TOG_2018} using initial and refined depth maps. Inaccurate depth values in the initial prediction result in an unnatural sharp background region. With our refined depth map, it is corrected and blurry.

\smallskip\noindent
\textbf{Analysis on mask quality}\quad 
We provide a visual comparison using different masks coupled with the same image and a numerical analysis with degraded masks in the appendix.
We show that our method can improve the depth quality as long as the given mask contains more accurate details than the original depth map.

\section{Conclusion}
Although depth maps are widely used in many practical applications, obtaining sharp and accurate depths from a single RGB image is highly challenging. In this paper, we presented the novel problem of mask-guided depth refinement and proposed a layered refinement approach that can be trained in a self-supervised fashion. Our method can significantly enhance initial depth maps quantitatively and qualitatively. We extensively validated our method by comparing it to mask-guided depth refinement baselines and existing automatic refinement methods. Furthermore, we verified that our method works well on real images with various masks and improves the results of downstream applications. 
We believe that our method can be potentially extended to other types of dense predictions such as normals and optical flow.
More results are provided in the appendix.

{
\smallskip\noindent
\textbf{Limitations}\quad
Since our method relies on a high-quality mask for refinement, its refinement performance is bounded by the mask quality. Although many auto-masking tools are available, capturing extremely fine-grained details may require manual work. Furthermore, as our method refines along mask boundaries, initially wrong depth values inside objects are likely to be left unaltered.
}

{\small
\bibliographystyle{ieee_fullname}

}

\begin{appendices}
\section{Potential Negative Societal Impact}
As our proposed method refines depth maps predicted by SIDE models, we do not expect it to have any direct negative societal impact. However, potentially, it can be used to generate more accurate 3D reconstructions of people, and if used in a malicious way, they could be reconstructed accurately in an unwanted way.

\section{Image Copyrights}
Comparison images in Fig.~\ref{fig:qual_hypersim} and Fig.~\ref{fig:supp_qual_hypersim} are results on the Hypersim dataset (CC-BY SA 3.0 License) \cite{Roberts_ICCV_2021}. Images with human subjects (identifiable and non-identifiable) in Fig.~\ref{fig:teaser}, \ref{fig:motivation}, \ref{fig:applications}, \ref{fig:supp_hole} and \ref{fig:supp_point_cloud} are from \texttt{unsplash} \cite{unsplash} or \texttt{pixabay} \cite{pixabay}, which are websites with freely licensed images that can be used for commercial and non-commercial purposes. The top image in Fig.~\ref{fig:qual_sky} was officially licensed by Adobe Stock \cite{AdobeStock} (from eranda - stock.adobe.com). Other generic images are from internal RGB-D datasets.

\section{Details of Training Data Generation}

\noindent
\textbf{Perturbations}\quad
During training, we apply random dilation and erosion operations on the composite depth map. First, a random number of iterations is selected from $U(1, 5)$ each for dilation ($k_d$) and erosion ($k_e$). Then, dilation or erosion with a $3\times 3$ kernel is applied $k_d$ or $k_e$ times with the following order: (i) dilation, erosion, erosion and dilation for $50\%$ of the time, and (ii) erosion, dilation, dilation and erosion the rest of the time. This makes sure that most thin structures and isolated regions are lost in the perturbed depth map. For the Gaussian blur, $50\%$ of the time, we use $\sigma\sim U(0, 1)$ for small amounts of blur, and the rest of the time, we use $\sigma\sim U(1, 5)$ for larger amounts of blur. For human hole perturbations, holes in the mask are detected using the  hierarchy computed by \texttt{cv2.findContours()}, and for each hole, a random value between the mean depth value inside the original hole and the mean depth value in the outer neighborhood of 10 pixels is assigned.

\begin{figure}
\centering
\includegraphics[width=\linewidth]{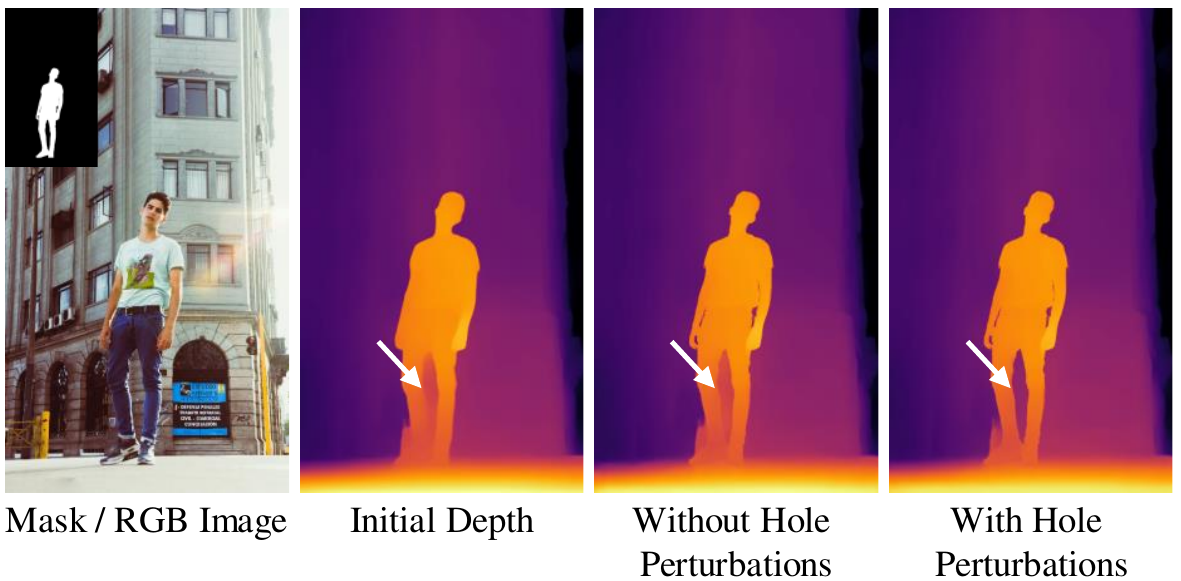}
\caption{Effect of human hole perturbation. By adding random human hole perturbations when generating the perturbed depth maps during training, our model can correct initially wrong values in large isolated background regions (\textit{holes}) in humans.}
\label{fig:supp_hole}
\end{figure}

\smallskip\noindent
\textbf{Effect of Human Hole Perturbation}\quad
We compare the refined depth results generated by a model trained \textit{without} human hole perturbations and our final model trained \textit{with} human hole perturbations (models in the last two rows in Table~\ref{table:ablation2}). As shown in Fig. \ref{fig:supp_hole}, the initial depth predicts wrong values for holes (isolated background regions) in humans. Without human hole perturbations, the model is able to refine smaller holes (between arm and body) but is incapable of correcting a larger hole (between the legs) as it has not seen such challenging cases during training. The hole perturbation scheme aims to mimic those cases by assigning a random value. This simple strategy enables the refinement model to correct larger holes, as shown in Fig. \ref{fig:supp_hole}.

\begin{figure*}
\centering
\includegraphics[width=\linewidth]{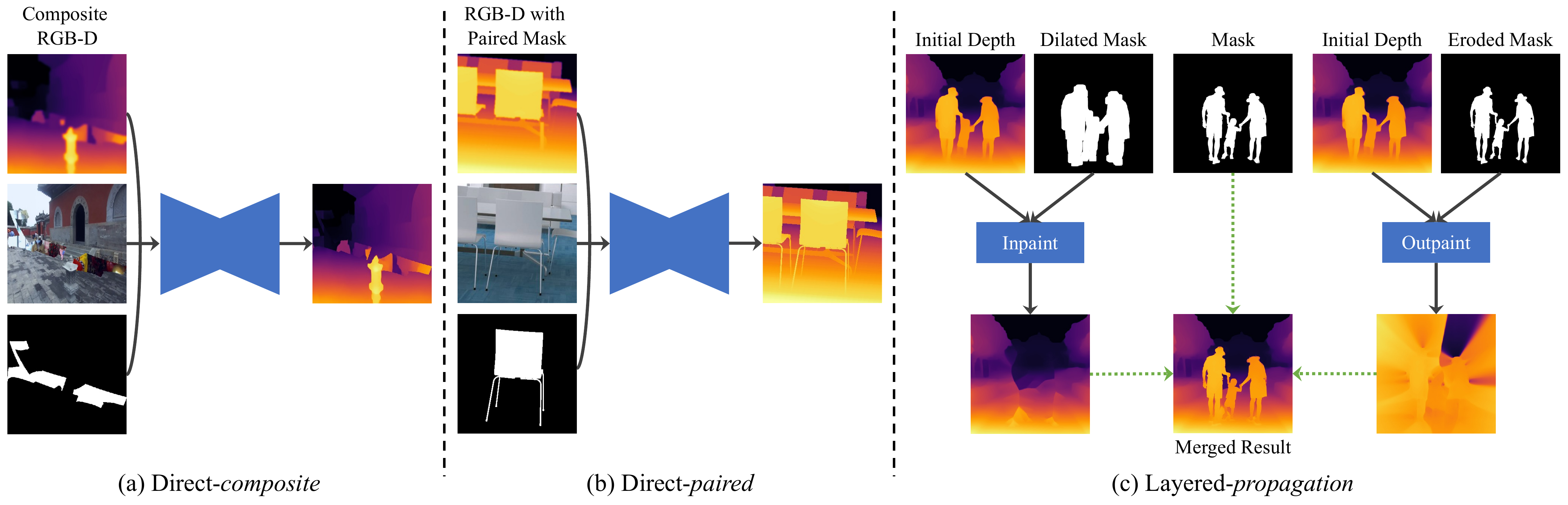}
\caption{Illustrations of baseline models used in our experiments.}
\label{fig:supp_baselines}
\end{figure*}

\begin{figure}
\centering
\includegraphics[width=\linewidth]{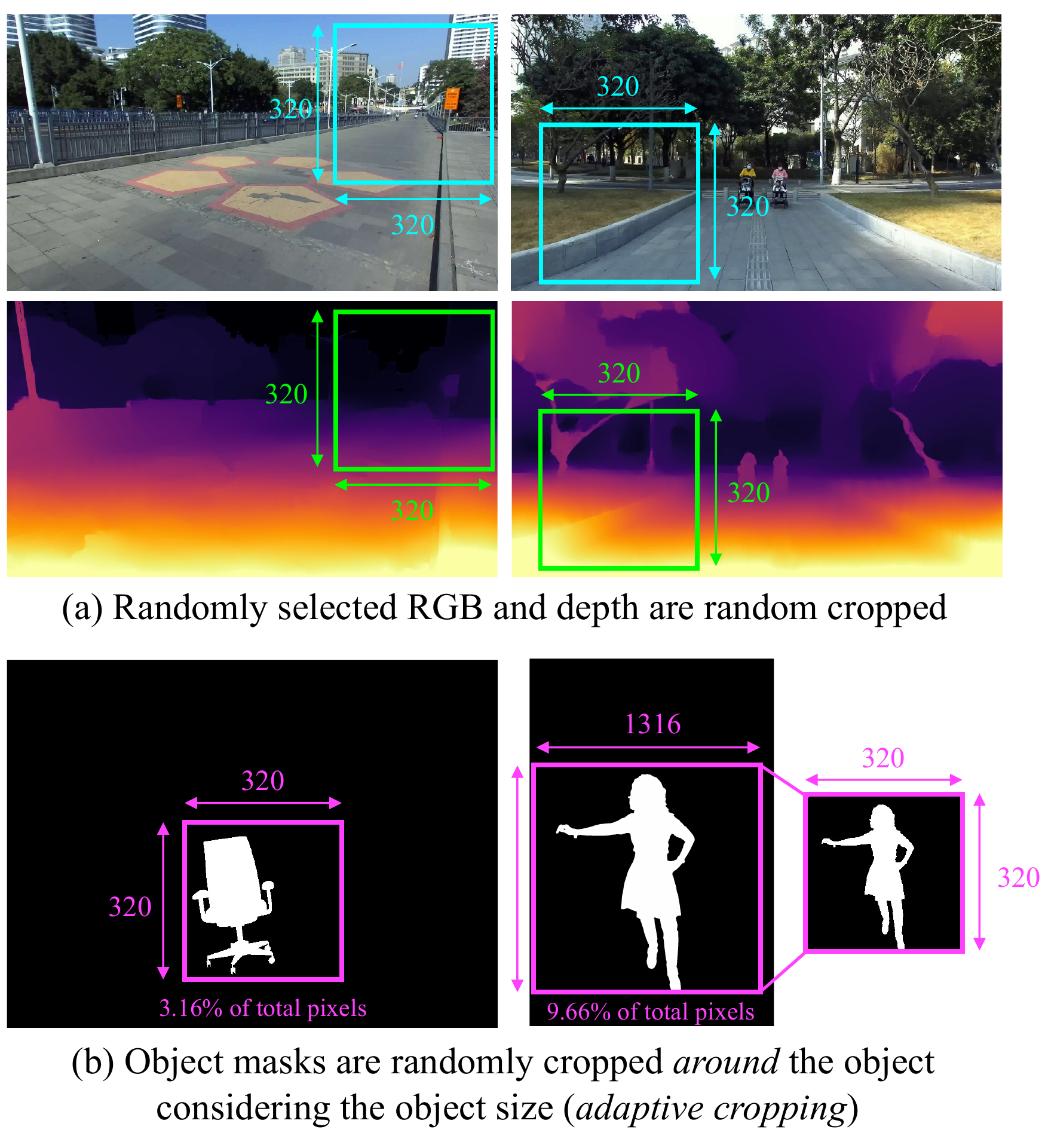}
\caption{RGB-D and mask cropping during training.}
\label{fig:supp_cropping}
\end{figure}

\smallskip\noindent
\textbf{Cropping}\quad
When cropping the mask for training, we filter out small objects by randomly picking objects that are comprised of at least 1\% of total pixels in an instance segmentation map. Furthermore, we adaptively crop \textit{around} the object depending on the object size to ensure that the masked region is sufficiently large as shown in Fig.~\ref{fig:supp_cropping}. If the object size is smaller than the training patch size ($320\times 320$), we randomly crop by the patch size at locations where the entire object is inside the patch. If the object size ($H\times W$) is bigger than the patch size, we crop by $p\times p$, where $p\sim U(s, 2s)$ and $s$ is $\max{(H, W)}$, at random locations where the entire object is inside the patch. Then, the cropped patch is resized to the training patch size so that it can be used for randomly compositing the RGB and depth map patches. Without this cropping scheme, the mask region often only contains parts of objects or no objects at all (if simply cropped at random locations). For stuff classes (e.g., sky), we crop with $p\sim U(H/2, H)$ at a random location.

\section{Details of Baseline Models}
In the main paper, we compared to four baseline models that perform mask-guided depth refinement: Direct-\textit{composite}, Direct-\textit{paired}, Layered-\textit{propagation} and Layered-\textit{ours}, described in Section~\ref{sec:comp_methods}. An illustration of the baselines is shown in Fig. \ref{fig:supp_baselines}. In Fig. \ref{fig:supp_baselines} (a), Direct-\textit{composite} predicts the refined output without layering by training on composite RGB-D inputs. Direct-\textit{paired} also refines without layering but is trained on a paired mask and RGB-D dataset \cite{Roberts_ICCV_2021} as shown in Fig. \ref{fig:supp_baselines} (b). We employ the same model architecture as the network shown in Fig.~\ref{fig:net_arch} for Direct-\textit{composite} and Direct-\textit{paired}. 

For Layered-\textit{propagation}, we run the propagation-based image completion algorithm \cite{Telea_JGT_2004} twice to obtain layered outputs, once with the dilated mask for inpainting and the second time with the eroded mask for outpainting as shown in Fig. \ref{fig:supp_baselines} (c). The two outputs are then merged based on the mask similar to our proposed 2-layer approach. For Layered-\textit{ours}, the same procedure as Fig. \ref{fig:supp_baselines} (c) is applied but we use our model after stage I training instead of \cite{Telea_JGT_2004} for inpainting/outpainting. For the layered baselines, dilation and erosion are necessary to correct the initially wrong values and their kernel sizes should be set \textit{heuristically} for each input depth to get the best results, unlike our proposed method that is able to automatically figure out the regions to inpaint/outpaint while refining inaccurate areas \textit{without} any heuristics.

\begin{figure*}
\centering
\includegraphics[width=\linewidth]{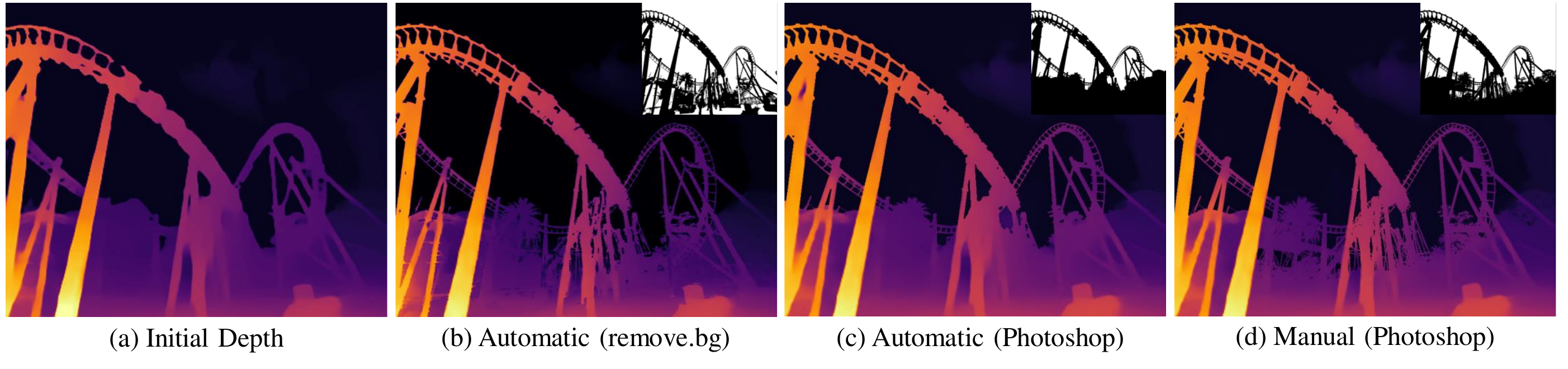}
\caption{Ablations on automatic and manual mask inputs.}
\label{fig:reb_auto_mask}
\end{figure*}

\begin{figure*}
\centering
\includegraphics[width=\linewidth]{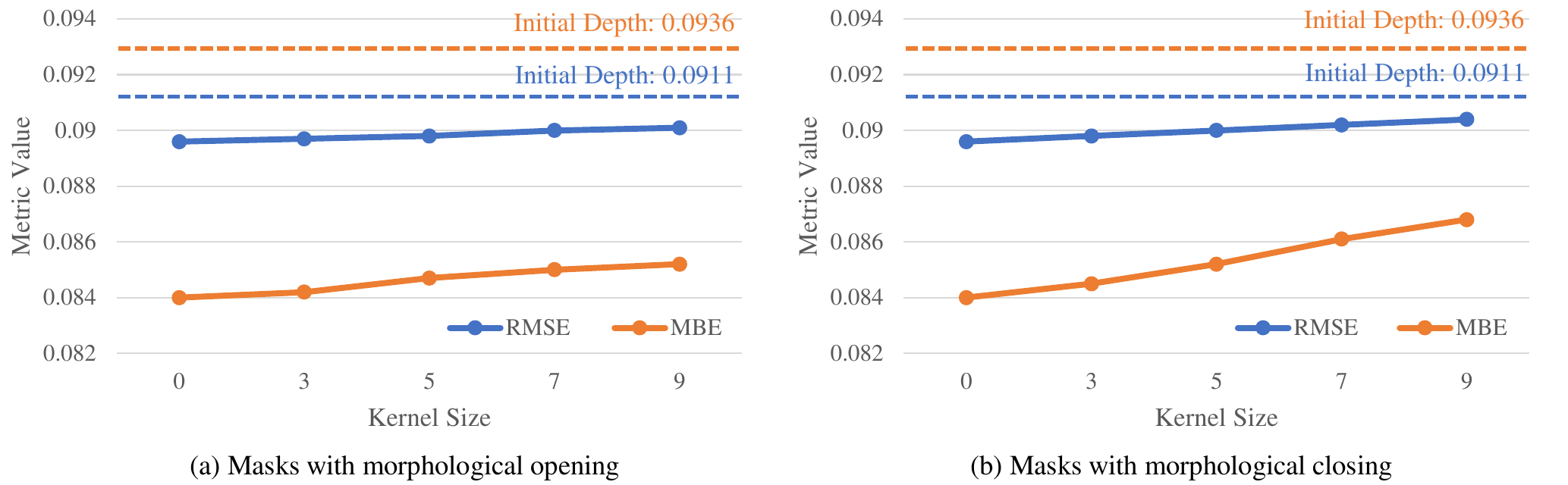}
\caption{Quantitative results with degraded masks.}
\label{fig:reb_mask_sensitivity}
\end{figure*}

\section{Analysis on Mask Quality}

As our method refines the initial depth map based on the input mask, its refinement performance is inevitably dependent on the mask quality. To analyze the effect of using different types of masks, in Fig.~\ref{fig:reb_auto_mask}, we show the refined outputs using three different masks generated using commercial masking tools: (i) automatically generated mask from \texttt{removebg}, (ii) automatically generated mask using \texttt{Photoshop}, and (iii) manually edited mask using \texttt{Photoshop}. As shown in Fig.~\ref{fig:reb_auto_mask}, using automatically generated masks already produces significantly enhanced results. With additional manual editing (Fig.~\ref{fig:reb_auto_mask} (d)), the depth map can be refined even further. In practical application scenarios, users can edit masks instead in order to edit depth maps, which would be easier and more intuitive.

For a numerical analysis on mask quality, we apply morphological opening and closing operations with kernel sizes $k\in\{3, 5, 7, 9\}$ on the ground truth instance segmentation maps from Hypersim \cite{Roberts_ICCV_2021} and measure the MBE and RMSE after refining the depth maps generated with DPT-Large \cite{Ranftl_ICCV_2021}. The results are plotted in Fig.~\ref{fig:reb_mask_sensitivity}, where $k=0$ denotes the case using the original ground truth segmentation maps and the dotted lines signify the average metric values of the initial depth maps. As shown in Fig.~\ref{fig:reb_mask_sensitivity}, the error values increase with more severe degradation as expected. However, they are still better than the initial depth.

\section{Inference Time}
For inference, it takes 16 ms for the initial depth prediction \cite{Ranftl_TPAMI_2020, Ranftl_ICCV_2021} and an additional 78 ms for our refinement method with an NVIDIA TITAN RTX GPU. Note that input images are resized to the spatial resolution used during training prior to entering the network for all methods, $384\times384$ for \cite{Ranftl_TPAMI_2020, Ranftl_ICCV_2021} and $320\times320$ for ours.

\section{More Visual Results}

\begin{figure*}
\centering
\includegraphics[width=\linewidth]{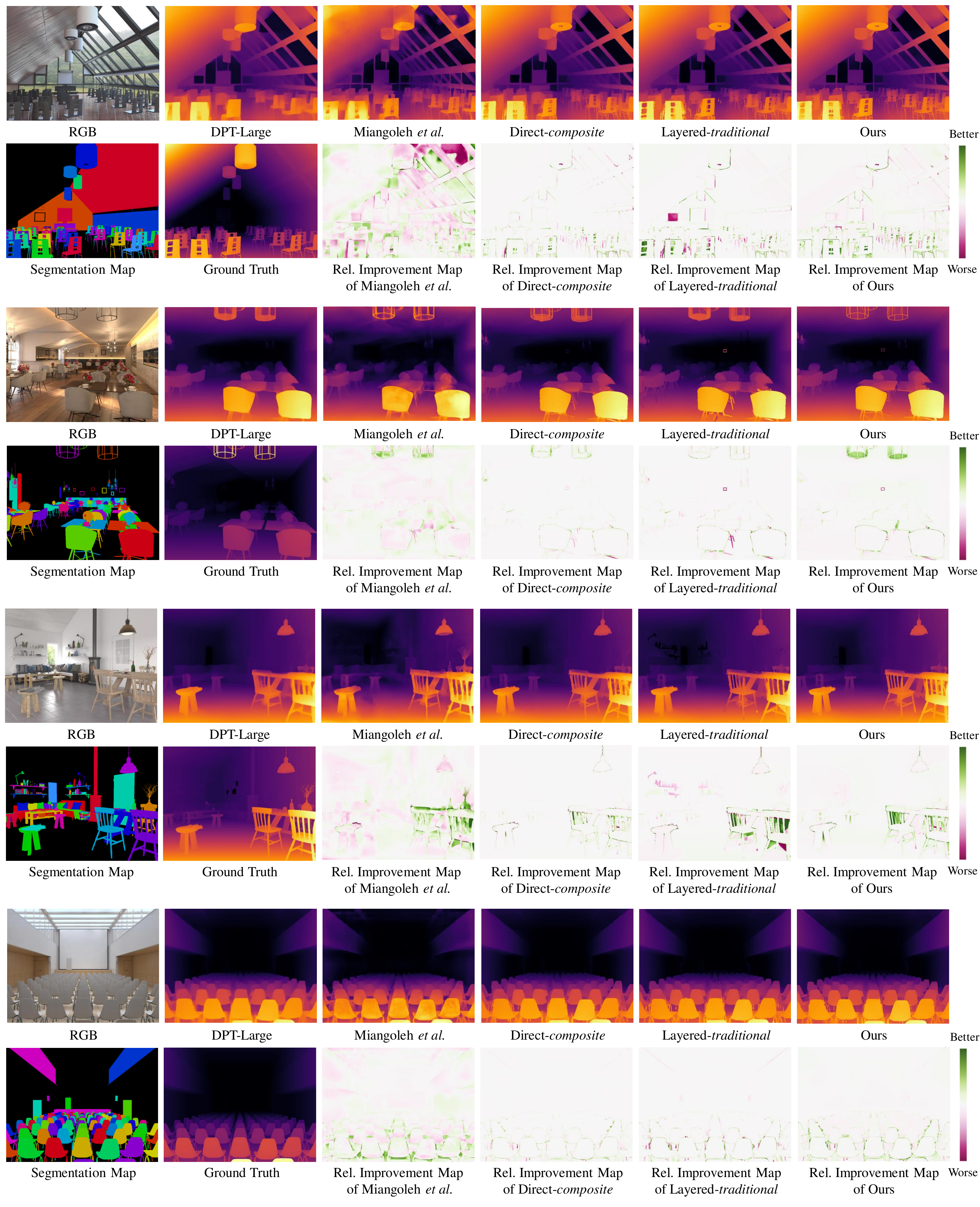}
\caption{Qualitative results on Hypersim \cite{Roberts_ICCV_2021}. The relative improvement maps visualize where the refinement method improved and worsened the initial depth estimation by DPT \cite{Ranftl_ICCV_2021}. Our method focuses on the edges and hole regions, accurately refining fine structures.}
\label{fig:supp_qual_hypersim}
\end{figure*}

\noindent
\textbf{More results on paired datasets}\quad
In Fig. \ref{fig:supp_qual_hypersim}, we provide more examples on Hypersim \cite{Roberts_ICCV_2021} along with the relative improvement maps visualizing where the refinement method improved and worsened the initial depth estimation in terms of absolute error. Miangoleh~\textit{et al.}'s method \cite{Miangoleh_CVPR_2021} often worsens homogeneous regions whereas our method mostly refines along the mask boundaries (edges and holes) and leaves other regions intact.

\begin{figure*}
\centering
\includegraphics[width=\linewidth]{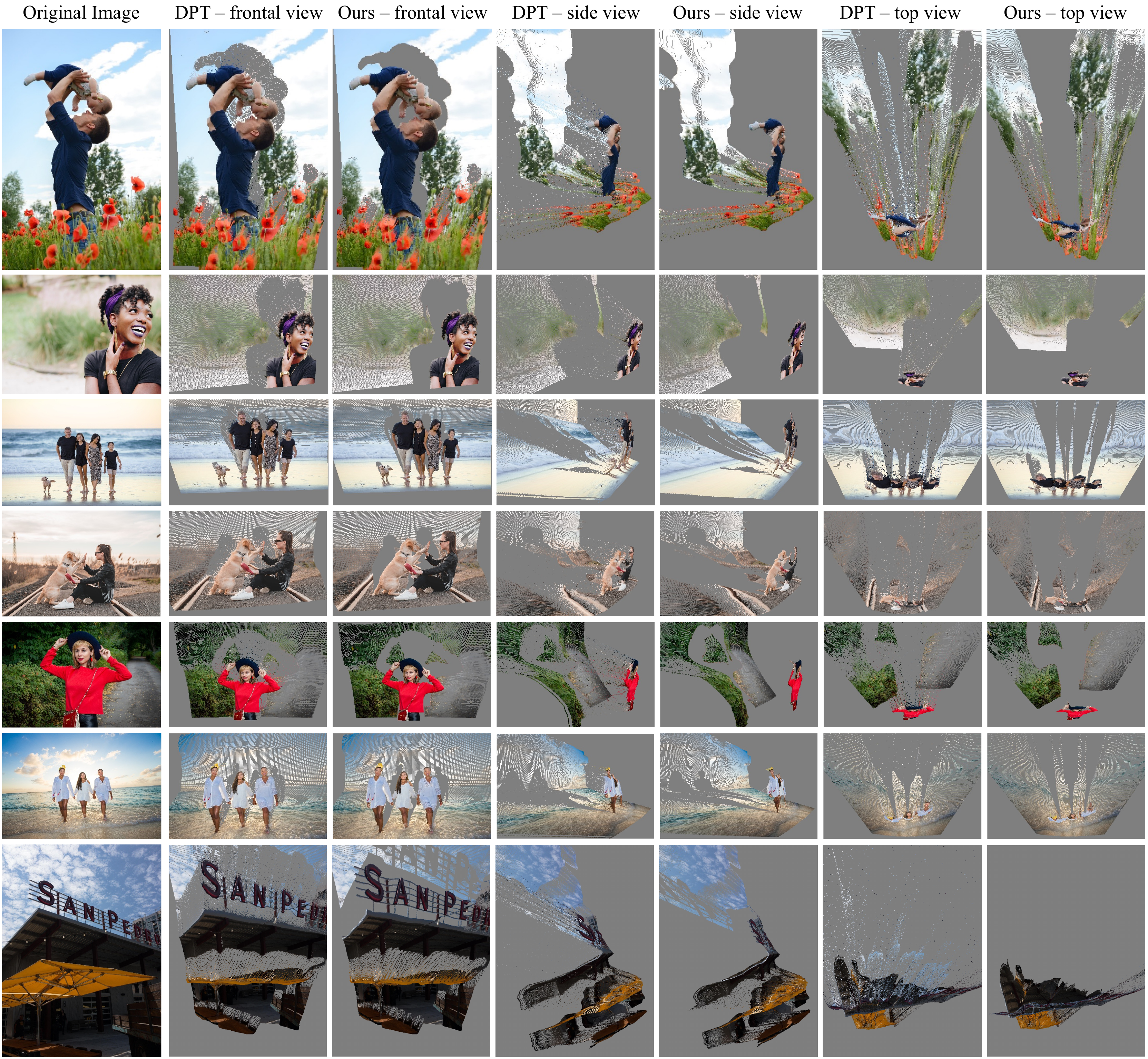}
\caption{Point cloud visualizations using the initial depth by DPT \cite{Ranftl_ICCV_2021} and refined depth by Ours. With the refined depth, there are less flying pixels and objects are more clearly cut in the frontal, side and top views of the scene.}
\label{fig:supp_point_cloud}
\end{figure*}

\smallskip\noindent
\textbf{More results using point clouds}\quad
In Fig. \ref{fig:supp_point_cloud}, we visualize the frontal, side and top views of the scene using point cloud representations. With our refined depth, objects are more clearly and accurately cut around the edges and hole regions, resulting in significantly less flying pixels. This can potentially benefit applications such as 3D photography \cite{Niklaus_TOG_2019, Shih_CVPR_2020}.

\smallskip\noindent
\textbf{More results in the wild}\quad
We provide additional results on real images as an html gallery on our project page\footnote[2]{\url{https://sooyekim.github.io/MaskDepth/}} for easier visual comparisons among the initial depth \cite{Ranftl_ICCV_2021}, Miangoleh~\textit{et al.}'s refinement method \cite{Miangoleh_CVPR_2021} and Ours.

\end{appendices}

\end{document}